\documentclass[a4paper,11pt]{article}
\usepackage{head,fullpage}     
\usepackage[pdftex]{graphicx}  
\usepackage{url}
\usepackage{datetime}

\usepackage[numbers]{natbib}

\usepackage[table,xcdraw]{xcolor}

\newdateformat{monthyeardate}{%
  \monthname[\THEMONTH] \THEYEAR}

\title{\textbf{Deep Reinforcement Learning in Quantitative Algorithmic Trading: A Review}}

\author{
        \textbf{Tidor-Vlad Pricope} \\
        The University of Edinburgh \\
        Informatics Forum, Edinburgh, UK, EH8 9AB \\
        \texttt{T.V.Pricope@sms.ed.ac.uk} \\
}

\begin{document}
\date{}
\maketitle

\begin{abstract}
        Algorithmic stock trading has become a staple in today’s financial market, the majority of trades being now fully automated. Deep Reinforcement Learning (DRL) agents proved to be to a force to be reckon with in many complex games like Chess and Go. We can look at the stock market historical price series and movements as a complex imperfect information environment in which we try to maximize return - profit and minimize risk. This paper reviews the progress made so far with deep reinforcement learning in the subdomain of AI in finance, more precisely, automated low-frequency quantitative stock trading. Many of the reviewed studies had only proof-of-concept ideals with experiments conducted in unrealistic settings and no real-time trading applications. For the majority of the works, despite all showing statistically significant improvements in performance compared to established baseline strategies, no decent profitability level was obtained. Furthermore, there is a lack of experimental testing in real-time, online trading platforms and a lack of meaningful comparisons between agents built on different types of DRL or human traders. We conclude that DRL in stock trading has showed huge applicability potential rivalling professional traders under strong assumptions, but the research is still in the very early stages of development.
\end{abstract}

\section{Introduction}

Stock trading is the buying and selling of a company's shares within the financial market. The goal is to optimize the return on the investment of the capital exploiting the volatility of the market through repetitive buy and sell orders. Profit is generated when one buys at a lower price than it sells on afterwards. Among the key challenges in the finance world, the increasing complexity and the dynamical property of the stock markets are the most notorious. Stiff trading strategies designed by experts in the field often fail to achieve profitable returns in all market conditions \cite{wu2020adaptive}. To address this challenge, algortihmic trading strategies with DRL are proposed.

Although automated low-frequency quantitative stock trading may seem like a fancy choice of words, it means something really simple. Automated refers to the fact that we are targeting algorithmic solutions, with very little to no human intervention. Quantitative stock trading refers to the fact that we want the trading agent developed to be able to take its decisions based on historical data, on quantitative analysis, that is based on statistical indicators and heuristics to dechiper patterns in the market and take appropiate actions \cite{wu2020adaptive}. Finally, low-frequency trading assures that the solutions that we seek resemble agents which can trade like a human being from 1 minute to a few days' timeframe. This choice is natural and does not narrow the literature down that much, as most high-frequency trading systems act at the level of pico-second or milli-seconds and the studies in this area are usually owned by high-profile companies which try to keep their research hidden. The hardware for such studies is also really limited and represents a reason for such a shortage of studies in high-frequency trading because it does not require only a few very good GPUs but also really small latency devices to connect to live datacenters that contain active stock prices. 

For the scope of this paper, there are further limitations that should be implied by the title and the abstract but nevertheless, they will be mentioned here. Our attention is focused on studies building an end-to-end trading agent using DRL as a core learning paradigm. We won’t discuss papers debating on forecasting stock prices or the direction of the market with deep learning. Although it is clear that one can develop a good trading agent if such a system would be successful, usually, stock prices are very volatile and can’t be predicted in such ways \cite{nabipour2020deep} leaving the job to a risk management system to do the work. From the same reason, we won’t mention papers that try to combine meta-heuristic algorithms (like Genetic Algorithms - GA \cite{goldberg2006genetic} or Particle Swarm Optimization - PSO \cite{kennedy1995particle}) with Reinforcement learning (RL) or multi-agent solutions to this problem. We also rule out the portfolio management part \cite{cooper2001portfolio} because it is entirely a different system that should be taken into consideration separately when constructing a trading engine \cite{soleymani2020financial, wang2019alphastock}. We focus on the actual agent - learner that makes the trades, the allocation of funds is not a priority to review in this paper.

It is no secret that the main theme of this paper is the applicability of AI in finance, further narrowed down by multiple research questions. As we will see, most of the work in this area is fairly new. The literature can answer the question quite well, as several benchmarks have been a staple for quite some time in algorithmic trading, like yearly return rate on a certain timeframe or cumulative wealth \cite{wang2019alphastock} or Sharpe ratio \cite{sharpe1994sharpe}. An issue that is usually problematic here relates to performance comparisons that can be inconclusive. Obviously, some papers test their proposed approaches on different environments and different timelines and may use a different metric for assessing performance, but a common ground can be found if similar settings are used.  Moreover, it is not futile to assume that many studies from the literature are kept private in this area, as it is a way to generate a lot of profit, therefore, state-of-the-art methods can hardly be called out; that's why we are going to divide this review paper according to the main idea of the methods used to solve the problem: critic-only RL, actor-only RL and actor-critic RL approaches, which is actually a well-defined categorization of RL in trading \cite{fischer2018reinforcement}.

There is an obvious question that this paper tries to answer, that being: \textit{can we make a successful trading agent which plays on the same timeframes as a regular human trader using the RL paradigm making use of deep neural networks?} We explore solutions presenting multiple papers in the current literature that propose approaches to the problem and actual concrete agents that we will be able to assess in order to define the progression on this field of study. We conclude that DRL has huge potential in being applying in algorithmic trading from minute to daily timeframes, several prototype systems being developed showing great success in particular markets. A secondary question that arises naturally from this topic is: \textit{can DRL lead to a super-human agent that can beat the market and outperform professional human traders?} The motivation for considering such a question is that, recently, it was showed that RL systems can outperform experts in conducting optimal control policies \cite{mnih2015human}, \cite{song2013fault} or can outperform even the best human individuals in complex games \cite{silver2018general}. We shall also try to respond to this question to some extent; we provide a partial answer supported by particular studies \cite{deng2016deep} where there have been DRL agents developed for real-time trading or for realistic back-testing \cite{huang2018financial} that can, in theory, allow for such a comparison. We infer that DRL systems can compete with professional traders with respect to the risk-adjusted return rates on short (15-minute) or long (daily) timeframes in particular hand-picked markets, although further research on this needs to be done in order to decipher the true extent of DRL power in this environment.

\section{Background}
In this section, we provide the necessary background for the reader to get a grasp of some basic concepts we shall recall throughout this review paper.

\subsection{Reinforcement Learning}

Reinforcement learning \cite{sutton2018reinforcement} is considered to be the third paradigm of learning, alongside supervised and unsupervised learning. In contrast with the other two, reinforcement learning is used to maximize the expected reward  when in an environment usually modelled as a Markov Decision Process (MDP). To be more precise, having as input a state representation of the environment and a pool of actions that can be made, we reach the next state with a certain numerical reward of applying a picked action. In order to approximate the expected reward from a state (or the \textbf{value} of a state), an action-value function $Q$ can be used. This takes as input a state and a possible action and outputs the expected $Q$ value \cite{watkins1992q} - the reward that we can expect from that state forward - the cumulative future rewards. In the literature, the action-value function is referred to as the \textbf{critic}, one might say it's because it "critiques" the action made in a state with a numerical value. However, RL can directly optimize the policy that the agent needs to take in order to maximize reward. If we view the policy as a probability distribution over the whole action space, then policy gradient (PG) methods \cite{silver2014deterministic} can be applied to optimize that distribution so that the agent chooses the action with the highest probability, amongst all possible actions, that gives the highest reward. Judging by the fact that this method directly controls to policy, in the literature, it is referred as the \textbf{actor}. RL employs these two approaches in the literature to form 3 main techniques: critic-only, actor-only, actor-critic learning. In the critic-only, we use only the action-value function to make decisions. We can take a greedy approach - always choosing the action that gives the max possible future reward, or we can choose to explore more. Note that the Q function can be approximated in many ways, but the most popular and successful approach is with a neural network (Deep Q Network - DQN \cite{mnih2015human} and its extensions \cite{van2016deep}). In the actor only approach, we model only the probability distribution over the state - the direct behaviour of the agent. In actor-critic RL, the actor outputs an action (according to a probability distribution) given a state and the critic (we can look at this as a feedback that "critiques" that action) outputs a $Q$ value (the future reward) of applying the chosen action in that given state. Popular, state-of-the-art algorithms like Advantage Actor Critic (A2C) \cite{zhang2020deep, mnih2016asynchronous}, Deep Deterministic Policy Gradient (DDPG) \cite{lillicrap2015continuous, xiong2018practical}, Proximal Policy Optimization (PPO) \cite{schulman2017proximal, liang2018adversarial} all follow this scheme.

\subsection{Technical analysis in stock trading: stock market terminologies}

The generally adopted stock market data is the sequence at regular intervals of time such as the price open, close, high, low, and volume \cite{wu2020adaptive}. It's a difficult task, even for Deep Neural Networks (DNN), to get a solid grasp on this data in this form as it holds a great level of noise, plus the non-stationary trait. Technical indicators were introduced to summarize markets' behaviour and make it easier to understand the patterns in it. They are heuristics or mathematical calculations based on historical price, volume that can provide (fig. 1) bullish signals (buy) or bearish signals (sell) all by themselves (although not accurate all the time). The most popular technical indicators that will be mentioned in the review as well are: Moving Average Convergence Divergence (MACD) \cite{baz2015dissecting}, Relative Strength Index (RSI) \cite{wilder1978new},  Average Directional Index (ADX) \cite{gurrib2018performance}, Commodity Channel Index (CCI) \cite{maitah2016commodity}, On-Balance Volume (OBV) \cite{tsang2009profitability} and moving average and exponential moving average \cite{hunter1986exponentially}. There are many more indicators and the encyclopedia \cite{colby1988encyclopedia} explains nicely each one of them.

\begin{center}
\begin{figure}[h]
\centering
\includegraphics[width=10cm]{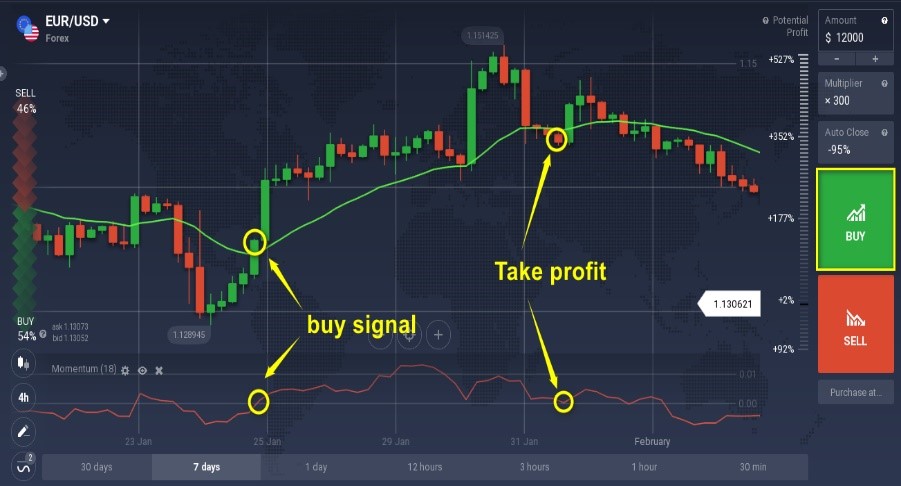}
\caption{Indicator showing buy and sell signals. Source: iqoption \cite{iqoption}}
\end{figure}
\end{center}
\vspace{-1cm}

The challenge that comes with applying technical indicators in algorithmic trading is the fact that there are so many of them and the majority do not work in specific situations - providing false signals. Moreover, many are correlated with each other - such dependencies need to be taken down as it can affect the behaviour of a learning algorithm because of the overwhelming obscure feature space. Solutions have been employed through deep learning - performing dimensionality reduction (through an autoencoder \cite{wang2014generalized}), feature selection and extraction (through Convolutional Neural Networks - CNN \cite{lecun1995convolutional} or Long Short Term Memory Networks - LSTM \cite{hochreiter1997long}), hence eliminating some noise in the raw market data.

These indicators alone can be a repertoire for a human trader and combinations between them can represent by themselves winning strategies for some stocks. That’s why there is actually a great amount of trading agents based on expert systems rule-based with great success using only technical analysis. There are also papers that present simple algorithms that are based only on data mining and the use of one indicator that can achieve over 32\% annual return \cite{kannan2010financial}.

\section{Literature Review}
In this section we review literature works regarding deep reinforcement learning in trading. There are key points we follow in this review, tied to the principal challenges that reinforcement learning faces when dealing with financial trading: data quality and availability (high quality data might not be free ot it might be limited for certain timeframes), the partial observability of the environment (there will always be a degree of un-observability in the financial market \cite{huang2018financial}) and the exploration/exploitation dilemma \cite{sutton2018reinforcement} (a great deal of exploration is usually required for a successful RL agent but this is not feasible in financial trading since random exploration would inevitably generate huge amount of transaction costs and loss of capital).

\subsection{Critic-only Deep Reinforcement Learning}
 \underline{\textit{Overview}}

Before getting to review the studies in this category, it is important to highlight the main limitations that such an approach suffers from. The critic method (represented mainly by the DQN and its improved variations) is the most published method in this area of research \cite{zhang2020deep} (compared to the others: actor-only and actor-critic). The main limitation is that a DQN aims to solve a discrete action space problem; so, in order to transform it to a continuous space (this can be needed as we will see in some reviewed papers), certain tricks are to be applied. Moreover, a DQN works well for discrete state representations, however, the prices of all the stocks are continuous-valued entities. Applying this to multiple stocks and assets is also a huge limitation as the state and action space grows exponentially \cite{xiong2018practical}. Moreover, the way that a reward function is defined in this context is crucial; critic-only approaches being very sensible to the reward signals it receives from the environment \cite{wu2020adaptive}.

 \underline{\textit{Related Works}}

\textbf{Application of Deep Reinforcement Learning on Automated Stock Trading} \cite{chen2019application}. Inspired by the success of state-of-the-art Deep Q Networks in tackling games like Atari \cite{mnih2013playing} and imperfect information games like Poker \cite{heinrich2016deep}, this paper tries to apply the same idea to develop a trading agent considering the trading environment as a game in which we try to maximize reward signaled by profit. A variant of the DQN learning system is tested by the authors: Deep Recurrent Q- Network (DRQN). The architecture is simple - they use a recurrent neural network at the basis of DQN which can process temporal sequences better and consider a second target network to further stabilize the process. The authors use S\&P500 ETF price history dataset to train and test the agent against baseline benchmarks: buy and hold strategy – which is used by human long-term investors and a random action-selected DQN trader. The data for training and testing is obtained through Yahoo Finance and it contains 19 years of daily closing prices, the first 5 being used for training and the rest for testing the agent. Therefore, a state representation is defined with a sequence of adjusted close price data over a sliding window of 20 days. Being a pure crittic approach, at any time, an agent has a discrete, finite set of actions - it has to decide between 3 options: buy a share, sell a share or do nothing. The reward is computed as the difference between the next day’s adjusted close and the current day’s adjusted close price, depending on which action was made.

As we can see in figure 2, the best agent, based on the DRQN system, is successful and outperforms the baseline models which gives hope for algorithmic trading. The annual expected return from these systems is around 22-23\%, which is not bad, but it is far off from some of the other reviewed agents in this paper that can get over 60\% annual return \cite{ huang2018financial}.

\begin{center}
\begin{figure}[h]
\centering
\includegraphics[width=10cm]{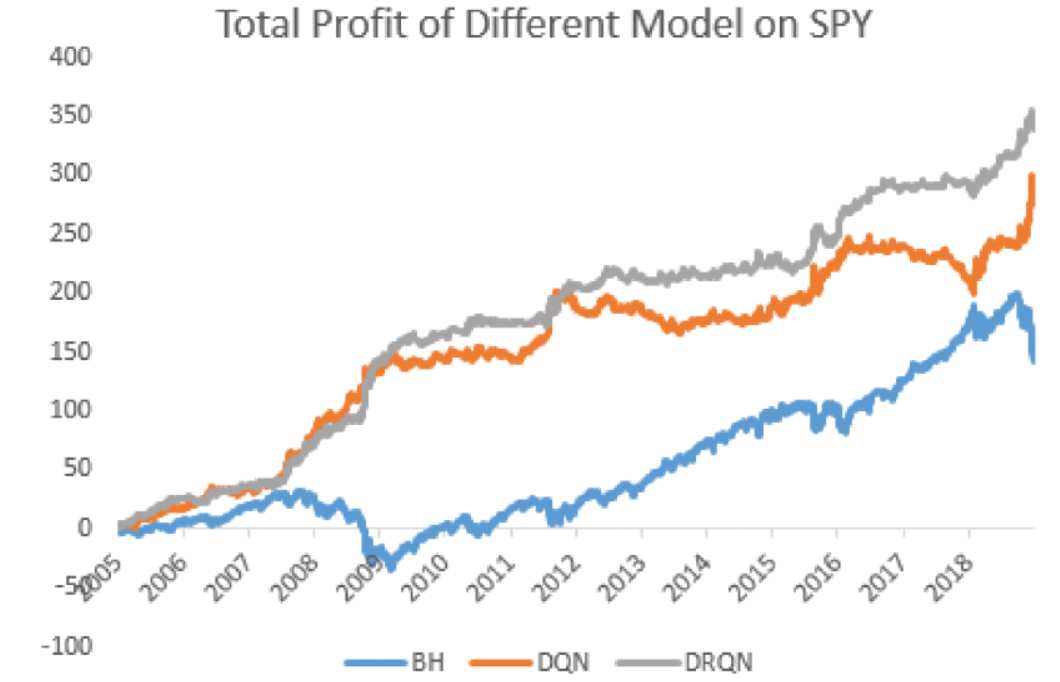}
\caption{Profit Curves of BH, DQN and DRQN on the SPY Test Dataset \cite{chen2019application}}
\end{figure}
\end{center}
\vspace{-1cm}

However, there are clear limitations on the study of this paper. First, only one stock is considered as a benchmark. That is not necessarily wrong but it does question the generality of the proposed approach to a larger portfolio, even from our own experiments doing algorithmic trading, some stocks can be very successful and some can lose all the allocated budget. Secondly, the reward function is really weak, it does not address risk as the other reviewed paper do, therefore it might be scary to let this model run on live data. A quick fix would be to use the Sharpe ratio as a reward measurement. Third, the agent is very constrained in what it can do - buying only one stock at a time is very limiting, more actions could be added to allow for more shares to be bought or sold. The feature representation is far from optimal, raw price data contains a lot of noise - technical analysis features should have been used to mitigate that. In terms of training/testing, it is concerning to us that the training data is much smaller in size and it might not be representative compared to the testing one - the 2006-2018 interval of time contains critical market states like the 2008 financial crisis in which the agent probably won't know how to handle. Furthermore, very basic opponents have been chosen for comparison purposes, even though the buy and hold strategy is objectively good and can be really profitable in real life, being actually the main strategy for investors, the paper lacks a direct comparison with other trading agents from the literature on the same stock. Nevertheless, we would categorize this paper as a solid pilot experiment into this wild field that can provide a proof-of-concept for the utility of deep reinforcement learning in trading.

A strongly related work with this is the study \textbf{Financial Trading as a Game: A Deep Reinforcement Learning Approach} \cite{huang2018financial} that also employs a DRQN-based agent. However, there are key differences:  it uses 16 features in total - raw price data (open, high, low, close, volume (OHLCV)) plus indicators; a substantially small replay memory is used during training; an action augmentation technique is also introduced to mitigate the need for random exploration by providing extra feedback signals for all actions to the agent. This enables the use of greedy policies over the course of learning and shows strong empirical performance compared to more commonly used $\epsilon$-greedy exploration. Moreover, they sample a longer sequence for recurrent neural network training.  Limitations still exist: the reward function (defined here as the portfolio log returns) is better as it normalizes the returns but still does not specifically address risk and the action space size is still very restricted (3 options).

The algorithm is applied on forex (FX) market from 2012 to 2017, 12 currency pairs, 15-minute timeframe with realistic transaction costs. The results are decent, the algorithm obtaining even 60\% annual return on some currencies, averaging about 10\% with all the results being positive, which is encouraging further research in this area. It is true that the authors haven't tested this algorithm in stock markets, only on FX, which would rule this paper out of scope for our review. However, considering the fact that this paper presents a better version of the previously-discussed algorithm, it is not wrong to assume it may be applied with a decent amount of success to the stock market as well. It's interesting here to also address the research question regarding the comparison with professional human traders. The environment for backtesting happens to have realistic setting in this paper and yet, the best result is a 60\% annual return, which is high. Nevertheless, it is impossible to just go with the numbers because we can't really know what an average professional trader or day-trader scores as profit rate in a year \cite{investopedia2021prof}, but we do know that 60\% a year in profits is much more than 7\% (the usual buy and hold annual return rate on investment for a stable growing stock index like S\&P 500) and would quickly generate more and more money if continuous reinvesting of funds is done and the algorithm is stable which is the main goal of traders and investors. Therefore, from this perspective, we can assume that, if consistent in trading, the algorithmic solution might be on par with human traders that make a living out of this.

There are a few studies \cite{wu2020adaptive, zhang2020deep} that employ critic-only deep reinforcement learning methods to compare with the other ones. In \textbf{Adaptive stock trading strategies with deep reinforcement learning methods} \cite{wu2020adaptive}, Gated Recurrent Units (GRUs) are introduced as a deep learning method to extract features from the stock market automatically for its Deep Reinforcement Q-Learning system (GDQN). The motivation for introducing this is the fact that stock market movements cannot reveal the patterns or features behind the dynamic states of the market. In order to achieve that, the authors present an architecture with an update gate and reset gate function for the neural network. The reset gate acts like a filter for the previous stock informationi, keeping only a portion of it, and the update gate decides whether the hidden state will be updated with the current information or not. This is a learnt way to tackle feature engineering for this problem, similar approaches with autoencoder architectures have been tried to achieve the same thing. To note,  Dropout \cite{srivastava2014dropout} is used with GRU in the Q-network.

\begin{center}
\begin{figure}[h]
\centering
\includegraphics[width=7cm]{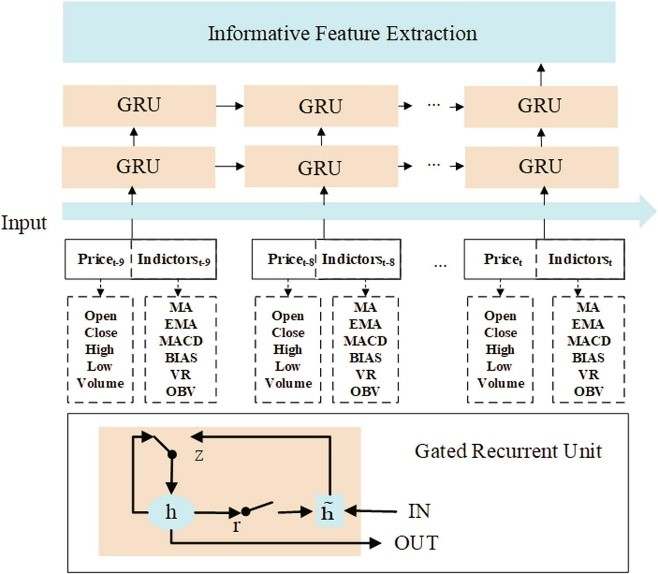}
\caption{Gated Recurrent Unit (GRU) architecture \cite{wu2020adaptive}}
\end{figure}
\end{center}
\vspace{-1cm}

 Another contribution of this paper is the use of a new reward function that better handles risk (something we repeatedly stated that it's missing before): the Sortino ratio (SR) \cite{mohan2016sortino}. The authors argue that this reward function would better fair in a high-volatility market because it measures negative volatility - it only factors in the negative deviation of a trading strategy’s returns from the mean. As state representation, the usual OHLCV is used plus some popular technical indicators like MACD, MA, EMA, OBV. The action space is again restricted to 1 share per transaction, 3 options being available: buy, sell or hold. Among the reviewed papers so far, this has the most promising setup; the daily stock data used stretched from early 2008 to the end of 2018, the first 8 years being used for training; it consisted of 15 US, UK and Chinese stocks, each one of them tested individually. The best result of the GDQN system was recorded on a Chinese stock: 171\% return and 1.79 SR. There was also recorded a SR of 3.49 and a return of 96.6\% on another Chinese stock, which is impressive. Just 2 out of 15 stocks were negative. Moreover, the system wins the comparison with the baseline (Turtle strategy \cite{faith2003original}) and it is showed to achieve more stable returns than a state-of-the-art direct reinforcement learning trading strategy \cite{deng2016deep}. GDQN is directly compared with the actor-critic Gated Deterministic Policy Gradient (GDPG) also introduced in this paper that we will review in the following sections. It is showed that GDPG is more stable than GDQN and provides slightly better results. Our criticism towards this study remains, like in the other ones, the limitation in what the agent can do - with only 1 share per transaction at max; in addition, transaction costs were not added in the equation, it would be interesting to see how the return rates would change if they would be factored in.

We will briefly mention \textbf{Deep Reinforcement Learning for Trading} \cite{zhang2020deep} as well, though their experiments focus on futures contracts and not stocks. It is very similar with \cite{chen2019application}, implementing a LSTM based Q-network, but making use of some technical indicators (RSI and MACD) to add to the feature space. As reward, it is important to mention that volatility of the market is incorporated in order to also consider the risk. This mechanism is used to scale up trade positions while volatility is low and do the opposite, otherwise. The authors show that show that the DQN system algorithms outperform baseline models (buy and hold, Sign(R) \cite{moskowitz2012time, lim2019enhancing}, MACD signal \cite{baz2015dissecting}) and deliver profits even under heavy transaction costs. This paper also tests an actor-only algorithm (PG) and the A2C algorithm to contrast with the baseline and with the DQN. We will get back to this paper in the next sections.

As we mentioned in the first paragraph of this section, the critic-only approach is the most published one in this field of financial trading; we had to remove some (older) studies \cite{tan2011stock, bertoluzzo2012testing, ritter2017machine} because they were out of the scope of this review paper - they weren't using deep Q-learning, only Q-learning with simple artificial neural networks (ANNs) or doing the feature extraction beforehand through other methods and not deep learning.
\subsection{Actor-only Deep Reinforcement Learning}
 \underline{\textit{Overview}}

With actor-only methods, because a policy is directly learnt, the action space can be generalized to be \textbf{continuous} - this is the main advantage of this approach: it learns a direct mapping (not necessarily discrete) of what to do in a particular state. A general concern we have discovered with this actor-only approach is the longer time to train that it takes to learn optimal policies, as also highlighted in \cite{zhang2020deep}. This happens because, in trading, learning needs a lot of samples to successfully train as otherwise, individual bad actions will be considered \textit{good} as long as the total rewards are great - we don't have a numerical value assigned to a particular state and action (like in the critic-only approach) that can tell us different through an expected outcome. However, this statement might be disputed \cite{fischer2018reinforcement}, as in some cases faster convergence of the learning process is possible.

 \underline{\textit{Related Works}}

\textbf{Deep Direct Reinforcement Learning for Financial Signal Representation and Trading} \cite{deng2016deep}. We briefly mentioned this study in the previous section (when comparing to the GDQN) as a state-of-the-art direct reinforcement learning trading strategy. This paper from 2017 specifically tries to answer the question of beating experienced human traders with a deep reinforcement learning based algorithmic-trader. This study is special because the authors claim that (at the time) it represented the first use of Deep Learning for \textbf{real-time} financial trading. Their proposed approach is a novel one. They remodel the structure of a typical Recurrent Deep Neural Network for simultaneous environment sensing and recurrent decision making that can work in an online environment. Like in the first paper reviewed, they use a Recurrent Neural Network for the RL part; however, they use Deep Learning for feature extraction combined with fuzzy learning concepts \cite{klir1988fuzzy, pal1994measuring} to reduce the uncertainty of the input data. They use fuzzy representations \cite{lin1991neural} (assign fuzzy linguist values to the input data) as part of the data processing step; after that, they use the deep learning for feature learning on the processed data and this is fed to the RL agent to trade. Another novelty that the authors introduce in this paper is a variant of Backpropagation through time (BTT) called task-aware BPTT to better handle the vanishing gradient problem in very deep architectures. To briefly explain this one, they use extra links from the output to each hidden layer during the backpropagation making more room for the gradient to flow. 

What is important here to note is the fact that the proposed solution that not use technical indicators, it utilizes the aforementioned system for raw data processing and feature learning to make educated guesses through the RL system to maximize profit and that is really impressive. we personally thought that technical indicators would greatly improve a system’s performance if we aim for the correct amount of un-correlated ones because of a simple reason: it adds more insight to the dataset and more features that can unravel the hidden patterns leading to a successful trading strategy. This does not mean that the performance using only raw data and clever data processing would beat current state-of-the-art models that make use of technical analysis but perhaps a comparison between their proposed system using and not using technical indicators would have been great to add. The experiments part is solid from a comparison point of view between established strategies in the literature and the proposed approaches; however, it lacks the testing on a diverse portfolio: only 3 contracts are chosen (2 commodities – sugar and silver and one stock-index future IF). It is actually addressed in the future work section that this framework should be adapted in the future to account for portfolio management. The comparisons (figure 4) on both the stock-index and commodity future contracts show that the proposed Direct Deep Reinforcement (DDR) system and its fuzzy extension (FDDR) generate significantly more total profit are much robust to different market conditions than Buy and Hold, sparse coding-inspired optimal training (SCOT) \cite{deng2015sparse} and DDR system without fuzzy representation for data processing. Moreover, on Sharpe ratio, the proposes systems recorded values over 9, which is incredibly high.

\begin{center}
\begin{figure}[ht]
\centering
\includegraphics[width=14cm]{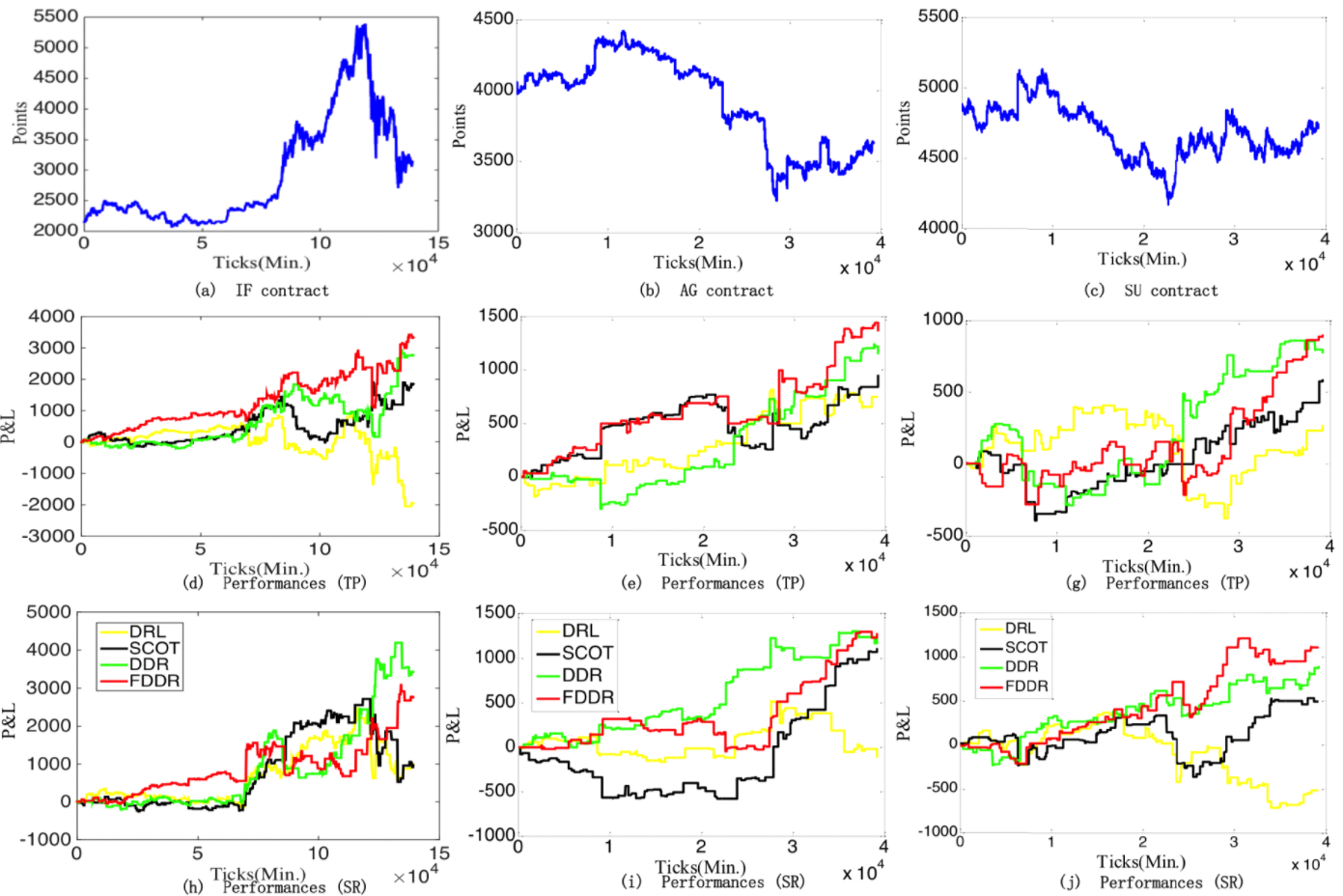}
\caption{Market behaviour (top), profit and loss (P\&L) curves of different trading systems with total profits (TP -middle) and moving Sharpe Ratio \cite{moody2001learning} (SR - bottom) \cite{deng2016deep}}
\end{figure}
\end{center}
\vspace{-1cm}

\begin{center}
\begin{figure}[ht]
\centering
\includegraphics[width=10cm]{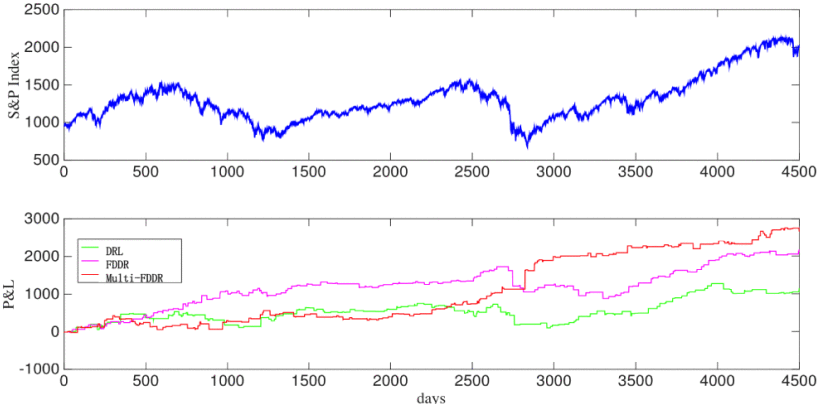}
\caption{S\&P 500 stock index P\%L curves on different trading systems  \cite{deng2016deep}}
\end{figure}
\end{center}
\vspace{-1cm}

The authors also tested the algorithms on stock indexes - like S\&P 500 (figure 5), where they can combine multi-market features to form a multi-FDDR agent. The stock data stretched from 1990 to 2015, the first 8 years being used for training and the transaction cost is set to 0.1\% of the index value. We can see (figure 5) that indeed the agents generate profits - although not that high considering the length of the testing period (about 18 years). 

One serious drawback of this study is the consistency of answering the main question that is opened and addressed in it. There is no comparison to human trading experts' performance in the experiments section, nor any discussion about this in the conclusion. Of course, the total profits displayed in the experiments do not come close to what a successful professional day trader or trading expert would achieve (because it does not come close to an actual annual wage), but this does not negate the premise. This is because the authors fail to mention the actual starting capital used in the experiments; therefore, we can only rely on the Sharpe ratio results for a conclusion. As we have mentioned before, the proposed approaches highlight remarkable high values with this metric: in the range 9-20. We need to be careful in calling this super-human performance, as investors usually consider a Sharpe ratio over 1 as good and over 3 as excellent \cite{investopedia}, but the paper does not mention if it uses the annualized Sharpe ratio or not. For more intuition on this, a ratio of 0.2-0.3 is in line with the broader market \cite{daytrading}. There is still a lot of skepticism behind these claims though, as the agents were tested on particular futures and commodities; one needs a broader portfolio to claim super-human success even though, at first, the results seem to be pointing in that direction. Nevertheless, the proposed agents are showed to generate a positive result in the long run which is an accomplishment by itself as there are some research studies \cite{barber2013behavior} arguing that many individual investors hold undiversified portfolios and trade actively, speculatively and to their own detriment. So, if we switch our research question to address and average trader, Deep Reinforcement learning approaches would probably be superior.

\textbf{Quantitative Trading on Stock Market Based on Deep Reinforcement Learning}  \cite{jia2019quantitative} is another study that explores a deep actor-only approach to quantitative trading. Its main contributions are a throughout analysis of the advantage of choosing a deep neural network (LSTM) compared to a fully connected one and the impact of some combinations of the technical indicators on the performance on the daily-data Chinese markets. This paper does prove that a deep approach is better and picking the right technical indicators does make a difference if we care about the return. The pure-performance results are mixed, the authors show that the proposed method can make decent profit in some stocks but it can also have oscillatory behaviour in others.

We have omitted impressive works like \textbf{Enhancing Time Series Momentum Strategies Using Deep Neural Networks} \cite{lim2019enhancing} in this section as it focuses more on the portfolio management part, plus it uses offline batch gradient ascent methods to directly optimize the objective function (maximizing the reward and minimizing the risk, Sharpe Ratio) which is fundamentally different from the standard actor-only RL where a distribution needs to be learnt to arrive at the final policy \cite{zhang2020deep}.

\subsection{Actor-critic Deep Reinforcement Learning}
 \underline{\textit{Overview}}

The third type of RL, actor-critic framework aims to simultaneously train two models at a time: the actor that learns how to make the agent respond in a given state and the critic - measuring how good the chosen action really was. This approach proved to be one of the most successful ones in the literature, state-of-the-art actor-critic DRL algorithms like PPO (in an actor-critic setting like it was presented in the original paper) or A2C solving complex environments \cite{schulman2017proximal}. However, in spite of this success, this method seems still unexplored; at least as of 2019 \cite{zhang2020deep}, it stayed among the least studied methods  in DRL for trading and as of current day, only a few new studies in this area have been published. Something that actor-critic algorithms like PPO do better than previous mentioned RL types is that it addresses well-known problems of applying RL to complex environments. One of it is that the training data that is generated during learning is itself dependent on the current policy so that means the data distributions over observations and rewards are constantly changing as the agent learns which is a major cause of instability. RL also suffers from a very high sensitivity to initialization or hyper-parameters: if the policy suffers a massive change (due to a high learning rate, for example) the agent can be pushed to a region of the search space where it collects the next batch of data over a very poor policy causing it to maybe never recover again.

 \underline{\textit{Related Works}}

\textbf{Deep Reinforcement Learning for Automated Stock Trading: An Ensemble Strategy} \cite{yang2020deep}. This paper proposed an ensemble system (figure 6) using deep reinforcement learning to further better the results in the literature with multiple stock trading experiments. They introduce an ensemble deep model that has 3 state-of-the-art DRL algorithms at core: PPO, A2C, DDPG. The authors state that the ensemble strategy makes the trading more robust and reliable to different market situations and can maximize return subject to risk constraints. The experiments are solid, being based on 30 stocks from Dow Jones portfolio. The features used to train the ensemble model are the available balance, adjusted close price, shares already owned plus some technical indicators: MACD, RSI, CCI and ADX.  The choice for the action space is solid as well: $\{ -k, ..., -1, 0, 1, ..., k \}$,  where $  k $ and $ -k $ presents the number of shares we can buy and sell, so in total there are 2k+1 possibility for one stock. This is later easily normalized to the continuous $[-1,1]$ action space for the main policy algorithms to work. It seems the authors do a better job at portfolio management problem with this approach - not being limited to buy only 1 share when in a great spot (like we saw in some of the previously reviewed papers). However, all the features from all the 30 stocks are merged, the aim being to also choose what stock to buy or sell – this is another part of portfolio management which is different and maybe should have been treated separately to the main trading algorithm.

\begin{center}
\begin{figure}[h]
\centering
\includegraphics[width=10cm]{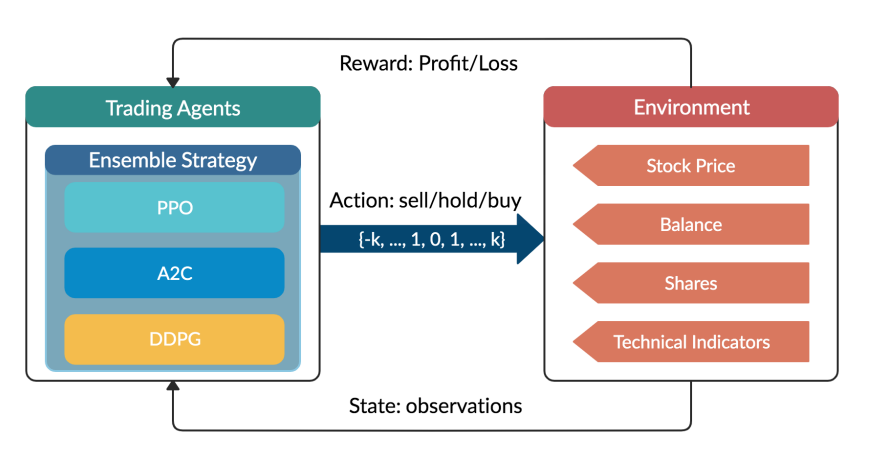}
\caption{Overview of the ensemble DRL strategy \cite{yang2020deep}}
\end{figure}
\end{center}
\vspace{-1cm}

\begin{center}
\begin{figure}[h]
\centering
\includegraphics[width=14cm]{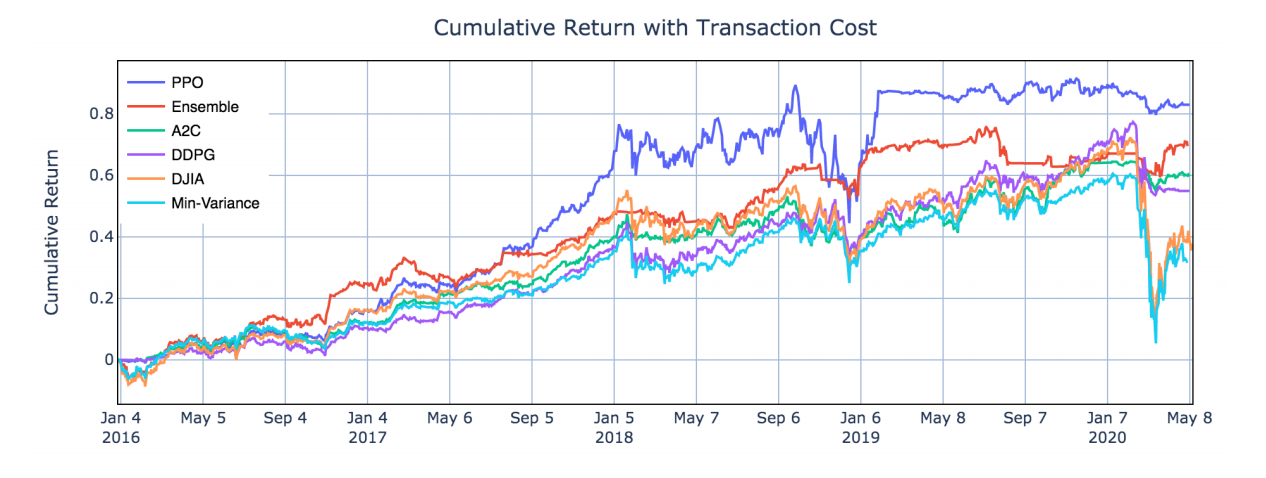}
\caption{Cumulative return curvess on different strategies on US 30 stocks index. Initital portfolio value: \$1m. \cite{yang2020deep}}
\end{figure}
\end{center}
\vspace{-1cm}

The way the ensemble strategy works is that a growing window training and a rolling window validation time period are chosen continuously so that we can pick the best performing algorithm over the validation set. This one is then used for the next fixed period of time as our sole agent. The motivation behind this is each trading agent is sensitive to different type of trends. One agent performs well in a bullish trend but acts bad in a bearish trend. Another agent is more adjusted to a volatile market.
This paper also discusses the impact of market crash or extreme instability to a trading agent. They employ the financial turbulence index that measures extreme asset price movements, using it together with a fixed threshold so that the agent sells everything and stops for making any trades if the markets seem to crash or be notoriously unstable. This is indeed a good idea; however, one can argue that it’s these periods of extreme volatility that a very good trading agent can exploit to make huge profits. 
The results of the experiments can be seen in figure 7. It seems the PPO algorithm still managed to provide slightly higher cumulative reward (at about 15\% annually); however, the Sharpe ratio which is the risk-adjusted reward is higher in case of the ensemble (which has 13\% return annually): $1.30$ compared to $1.10$. The proposed approach also beats the Dow Jones Industrial average index which is a staple benchmark for this portfolio and the min-variance portfolio allocation.

We personally think that a single model for a stock approach in this case will lead to much higher returns, instead of trying to build a master model that can handle all 30+ stocks at once. However, this still provides incredible results for the defined scope – handling portfolio management and the stock trading strategy simultaneously. The return is much greater than what you can expect through a buy and hold strategy (around 7\%) and the algorithm seems stable enough during extreme market conditions (the March 2020 stock market crash).

A strongly related paper with \cite{yang2020deep} is \textbf{Practical Deep Reinforcement Learning Approach for Stock Trading} \cite{xiong2018practical}. This paper only investigates the DDPG algorithm (from the previously reviewed ensemble) under the same conditions and with the same evaluation metrics but without any technical indicators as input. There are other minor differences: historical daily prices from 2009 to 2018 to train the agent and test the performance; Data from early 2009 to late 2014 are used for training, and the data from 2015 is used for validation; we test our agent’s performance on trading data, which is from early 2016 to late 2018. The results are pretty good: 25.87\% annualized return on this US 30 (Dow Jones) index, more than 15.93\% obtained through the Min-Variance method and 16.40\% the industrial average. The comparisons on Sharpe ratios only are also favorable (1.79 against 1.45 and 1.27) which makes it more robust than the others in balancing risk and return.

\textbf{Stock Trading Bot Using Deep Reinforcement Learning} \cite{azhikodan2019stock} is an especially interesting work because of the fact that it combines a deep reinforcement learning approach with sentiment analysis \cite{prabowo2009sentiment} (external information from news outlets) and proves that the proposed approach can learn the tricks of stock trading. Such a study is definitely important because many people are skeptical of the effectiveness of a pure quantitative learning algorithm in trading. Although the most important part of their system is the RL agent, this extra component that tries to predict future movements in the markets through sentiment analysis on financial news is an extra feature to the Deep Reinforcement Learning model. We can also view it as an extra precaution, an important signal tool in case of critical news or as a fail-safe system in case something major happens in the world that the pure quantitative trader is not aware of, though this is not what the authors directly had in mind.

The proposed approach in this paper is fairly simple at core: they use DDPG for the reinforcement learning agent and a recurrent convolutional neural network (RCNN) for classification of news sentiment. For the supervised learning part, the authors argue that a pure recurrent neural network (RNN) approach here would fail because it won’t determine discriminative phrases in a text, the advantage with convolutions being the max-pooling layers that come afterwards that can extract a good representation for the input. However, the recurring part is still needed as it provides contextual information to a greater extent. Hence, RCNN is used for this task - together with word embeddings, of course, for abstract text representation. Apart from the text embeddings, the change in stock price is also added as input and the network predicts a binary value $(0/1)$ - whether the price will go up (or down) in the following few days, this bit being merged with the information state for the RL agent.

\begin{center}
\begin{figure}[h]
\centering
\includegraphics[width=10cm]{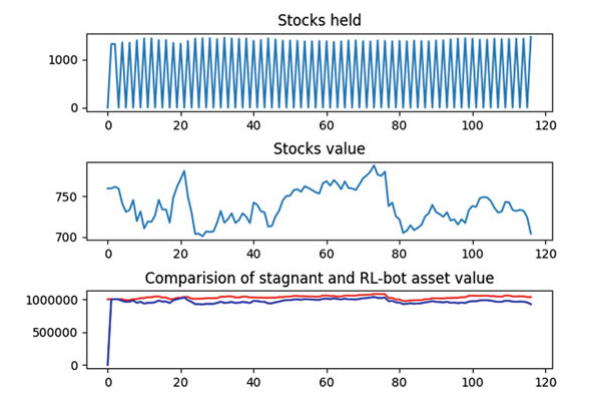}
\caption{Training over 5 months with NASDAQ-GOOGL stock. Initial budget: \$1m. The red line indicates the agent’s assets, and the blue line indicates the value of the stagnant stock \cite{azhikodan2019stock}}
\end{figure}
\end{center}
\vspace{-1cm}

The experiments involved market data on a daily timeframe for one stock, so really not on a large scale, but let’s not forget that the aim for this paper is more on the side of a proof of concept. The supervised learning system for sentiment analysis (RCNN) needed to be trained beforehand in order to accurately provide the extra feature for a possible future price movement according to the news outlets. The authors used almost 96k news headlines for training and around 31.5k for testing, in total, from 3.3k companies. The network is composed of embedding + Convolution + LSTM + Output with around 26M total model parameters. The results acquired are satisfactory: 96.88\% accuracy on test. However, information about the prior probability of the dataset used in missing, there should be a mention about the distribution to fully claim that such an accuracy is indeed good. Even though the experiments don’t show an overwhelming increase in profits (figure 8), it does prove that the agent does learn basic strategies: it does buy and sell continuously, it prefers to hold when there is a continuous decrease in price and it always maintains a higher value than the stagnant stock value. The authors also experimented with the reward function, trying 3 variants: difference between RL-agent asset value and stagnant asset value, difference between the cost at which the stocks are sold and the cost at which the stocks were bought, a binary reward representing if the action was profitable or not. The first 2 reward functions failed, the agent did not manage to learn this way, however, the last one managed to be the most successful and make the RL system work. Obviously, this is not the best reward system, it does not account for risk at all but it will sometimes work for the sole purpose of maximizing the gain.

There are two more work we have to include in this section. 

First, the GDPG system \cite{wu2020adaptive} we briefly mentioned in the critic-only RL section as a reference for comparison. The authors introduce GDPG as an actor-critic strategy that combines the Q-Network from the GDQN system with a policy network. It is showed (table 1) that this achieves more stable risk-adjusted returns and outperforms the baseline Turtle trading strategy. To put things into perspective, the turtle trading system is a well-known trend following strategy that was originally taught by Richard Dennis back in 1979. It was really successful back then and researchers use this system nowadays as a valid benchmark when comparing trading performances. The basic strategy is to buy futures on a 20-day high (breakout) and sell on a 20-day low, although the full set of rules is more intricate.

\begin{center}
\begin{figure}[h]
\centering
\includegraphics[width=10cm]{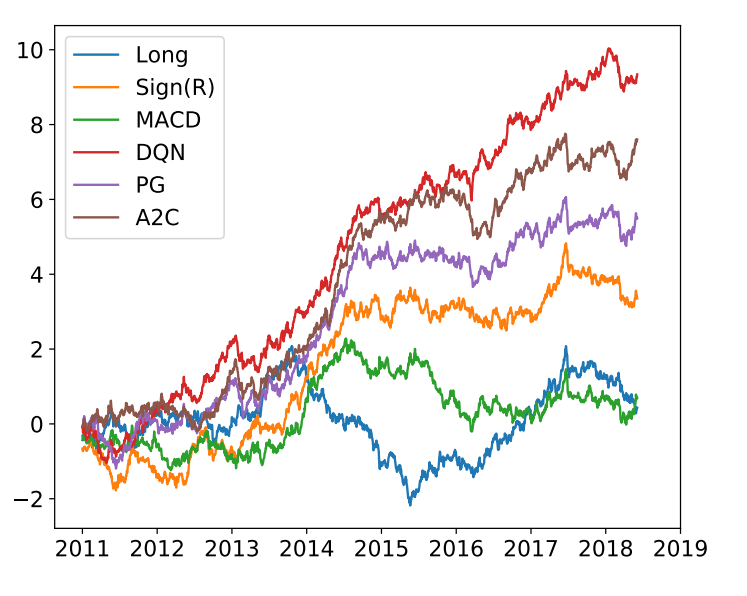}
\caption{Cumulative trade returns. In order: commodity, equity index and fixed income, FX and the portfolio of using all contracts \cite{azhikodan2019stock}}
\end{figure}
\end{center}
\vspace{-1cm}

\begin{center}
\begin{table}[]
\centering
\begin{tabular}{|c|cc|cc|cc|}
\multicolumn{1}{l}{}                   & \multicolumn{2}{c}{GDQN}                                                   & \multicolumn{2}{c}{GDPG}                                                   & \multicolumn{2}{c}{Turtle}                                                 \\
{\textbf{Symbol}} & { \textbf{SR}} & {\textbf{R(\%)}} & {\textbf{SR}} & { \textbf{R(\%)}} & { \textbf{SR}} & {\textbf{R(\%)}} \\
{AAPL}            & {1.02}        & { 77.7}           & { 1.30}        & {82.0}              & {1.49}        & { 69.5}           \\
{GE}              & {-0.13}       & {-10.8}          & {-0.22}       & { -6.39}             & { -0.64}       & {-17.0}          \\
{AXP}             & {0.39}        & { 20.0}           & { 0.51}        & {24.3}           & {0.67}        & { 25.6}              \\
{CSCO}            & { 0.31}        & { 20.6}              & { 0.57}        & {13.6}           & {0.12}        & {-1.41}          \\
{IBM}             & { 0.07}        & { 4.63}              & {0.05}        & {2.55}           & {-0.29}       & { -11.7}         
\end{tabular}
\vspace{1cm}
\caption{Comparison between GDQN, GDPG, Turtle strategy in some U.S. stocks over a 3-year period (2016-2019) \cite{wu2020adaptive} }
\end{table}
\vspace{-1cm}
\end{center}

Secondly, there is \cite{zhang2020deep} that we said we will get back to in section 3.1. This study directly compares the A2C algorithm with an actor-only approach (PG) and the critic-only approach we discussed in section 3.1. A graph with their experimental results can be seen in figure 9; it seems the A2C comes in second place on the whole portfolio return, this is explained by the authors saying that A2C generates larger turnovers, leading to smaller average returns per turnover. The experiments also showed that the A2C algorithm can monetize on large
moves without changing positions and also deal with markets that are more \textit{mean-reverting} \cite{zhang2020deep}.

\section{Summary \& Conclusion}
Deep Reinforcement Learning is a growing field of interest in the financial world. All the papers that were discussed here had their own experimental settings, assumptions and constrains. This makes it hard to compare one against another especially when the scope of the study is different between them, a direct comparison is not feasible in this scenario. Nevertheless, comparisons between different types of DRL methods were possible \cite{chen2019application, huang2018financial, zhang2020deep, deng2016deep} in the context of same study but also, limited, in the context of different studies as we saw with \cite{wu2020adaptive} and \cite{deng2016deep}. We have observed that DRL can be a powerful tool in quantitative low-frequency trading, a few studies with more realistic setups \cite{huang2018financial, wu2020adaptive, deng2016deep} obtaining over 20\% annual return with risk-aware metrics also being used. We have come to the conclusion that such agents can rival human traders in specific markets, but further research should be conducted in this area.

There is academic research \cite{thomas2013evaluation, clare2013evaluation} (granted not that recent compared to the papers reviwed here) that claims the market cannot be predicted and taken advantage of, supporting the hypothesis with data and experiments. With this paper, we have also disputed such claims. It is true that some markets cannot be predicted under some experimental setups, such as equities traded on a daily basis, as the authors of the mentioned studies showed, but this does not mean no market can be predicted in any setting - the above reviewed literature backs that up.

A lot of reviewed papers focused on trading on a daily timeframe. There is a fundamental problem with this that is barely brought up in the discussion section of the reviewed studies. Over longer time scales, such as days and weeks, market activity can be a direct result of complex interactions between political news, legal rulings, public sentiment, social hype, business decisions, etc. Large trades by institutional investors can also have a big impact. These are not aspects that can be modeled or exploited by quantitative algorithms (e.g. the well known March 2020 market crash caused by COVID-19). However, if we zoom into the market activity for a single hour, minute, or second, we can often see patterns. It is believed that these patterns are a result of algorithmic rule-based trading that we can exploit with an intelligent system.

Another concern that we have, especially for the testing stage of the proposed solutions to the problem discussed here, is that many of the papers consider years on end as testing intervals of time to assess the performance of a learning algorithm. For proof-of-concept purposes, this is not a problem; however, in a realistic setting, one would not deploy such an agent in the wild, the stock market is dynamic and needs constant retraining as covariance shifts. Trading/testing sessions of a fixed, relatively small length (2-3 months) would be much more relevant because we make use of more data (the latest data units to be precise) and the learning algorithm would be much more caught-up with the market in this way.

 Future possible directions of research include:

 \begin{itemize}
  \item More extensive experiments on live-trading platforms rather than backtesting or very limited real-time trading.
  \item Direct comparisons between DRL trading agents with human traders. For example, having a professional day trader run against an algorithmic DRL agent over the same market and interval of time in a controller experiment to observe which one would get higher returns.
  \item More comparisons among state-of-the-art DRL approaches under similar conditions and data sources.
  \item More research into the behaviour of DRL agents under critical market conditions (stock market crashes).
\end{itemize}

Although researching these points would greatly advance the field of AI in finance, there are non-trivial problems to be solved along the way, that even studies we have reviewed in this paper, which claim that they have a realistic setting, fall short of addressing some of them. 

First, if we are relying on machine learning (pattern matching/DRL) for our trading agent, we can't hope to make good predictions on higher time scales, as we have explained above. However, if we go too low (timeframewise), the transaction fees might overcome the actual small gains of relative high frequency trading. Therefore, picking the perect timescale is not a trivial problem and should be addressed more. 

Secondly, a really good model (according to backtesting) is not enough to secure profits, as most of the papers we have reviewed seemed to suggest. As we saw, the final step in finance research is often backtesting on historical data. If the model does well, the researchers declare success, usually not addressing the fact that the model would probably be questionable in a production environment. This does not mean that there is no research there on real-trading but if their new algorithm performed as well in the real-world, they certainly would not publish a paper about it and give away their edge. To expand on this matter, on many other fields (computer vision, voice recognition), machine learning, train-test performance directly correlates with live performance as the data distribution does not change significantly over time, we can be pretty sure that a model performing well on the test set also does so in production. In trading, our training, backtesting, and live environments can become so different that we can't really make any guarantees.

In third, the backtesting should be generally improved. It should simulate a live trading environment really well, meaning latencies, non-standard order types, exchange commission structures and slippages.  And even so, latencies in the real world may be stable during low-activity periods and spike during high-activity periods, something that would be incredibly difficult to simulate. For example, our model can be really good but the results are poor in a live enviornment becaues the connection (latency) is high and we won't be able to observe new data fast enough before the competition processes it already. It should also be bug-free and handle \textbf{edge cases} really well: in simulation everything works perfectly, but in the real world we run into \textbf{API issues}, \textbf{request throttling}, and \textbf{random order rejections} during busy periods. Therefore, no matter how good the backtesting seems to be, it is still fundamentally different from a live environment, but we can still try to amend some of these issues.

\bibliographystyle{unsrt}

\small
\bibliography{main}       

\end{document}